\documentclass[10pt,twocolumn,letterpaper]{article}
\usepackage{cvpr}
\usepackage[dvipsnames]{xcolor}
\newcommand{\red}[1]{{\color{red}#1}}

\definecolor{cvprblue}{rgb}{0.21,0.49,0.74}
\usepackage[pagebackref,breaklinks,colorlinks,citecolor=cvprblue]{hyperref}
\usepackage{algorithm}
\usepackage{algpseudocode}
\usepackage{amsfonts}
\usepackage{graphicx}
\usepackage{float}
\usepackage{colortbl}
\usepackage{mathtools}
\usepackage{multicol}
\usepackage{multirow}
\usepackage{ulem}
\usepackage{booktabs}

\newcommand{\delong}{\textcolor[rgb]{0.0,0.0,0.0}}

\newcommand{\green}{\textcolor[rgb]{0.0,1.0,0.0}}
\usepackage[utf8]{inputenc}
\usepackage[absolute,overlay]{textpos} 
\usepackage{lipsum}    

\title{Learning Implicit Features with Flow Infused Attention \\ for Realistic Virtual Try-On}

\begin{document}

\author{
    Delong Zhang$^{1*\dagger}$  \hspace{5mm}
    Qiwei Huang$^{2\dagger}$  \hspace{5mm}
    Yuanliu Liu$^{2}$  \hspace{5mm}
    Yang Sun$^{2}$  \hspace{5mm} \\
    Wei-Shi Zheng$^{1}$  \hspace{5mm}
    Pengfei Xiong$^{2\ddagger}$  \hspace{5mm}
    Wei Zhang$^{2}$  \hspace{5mm} \\
    \vspace{5mm}
    \textsuperscript{1}Sun Yat-sen University \hspace{5mm}
    \textsuperscript{2}Shopee \hspace{5mm}
    \vspace{-6mm}
}

\twocolumn[{
\maketitle
\captionsetup{type=figure}
\centering
\includegraphics[width=0.97\textwidth]{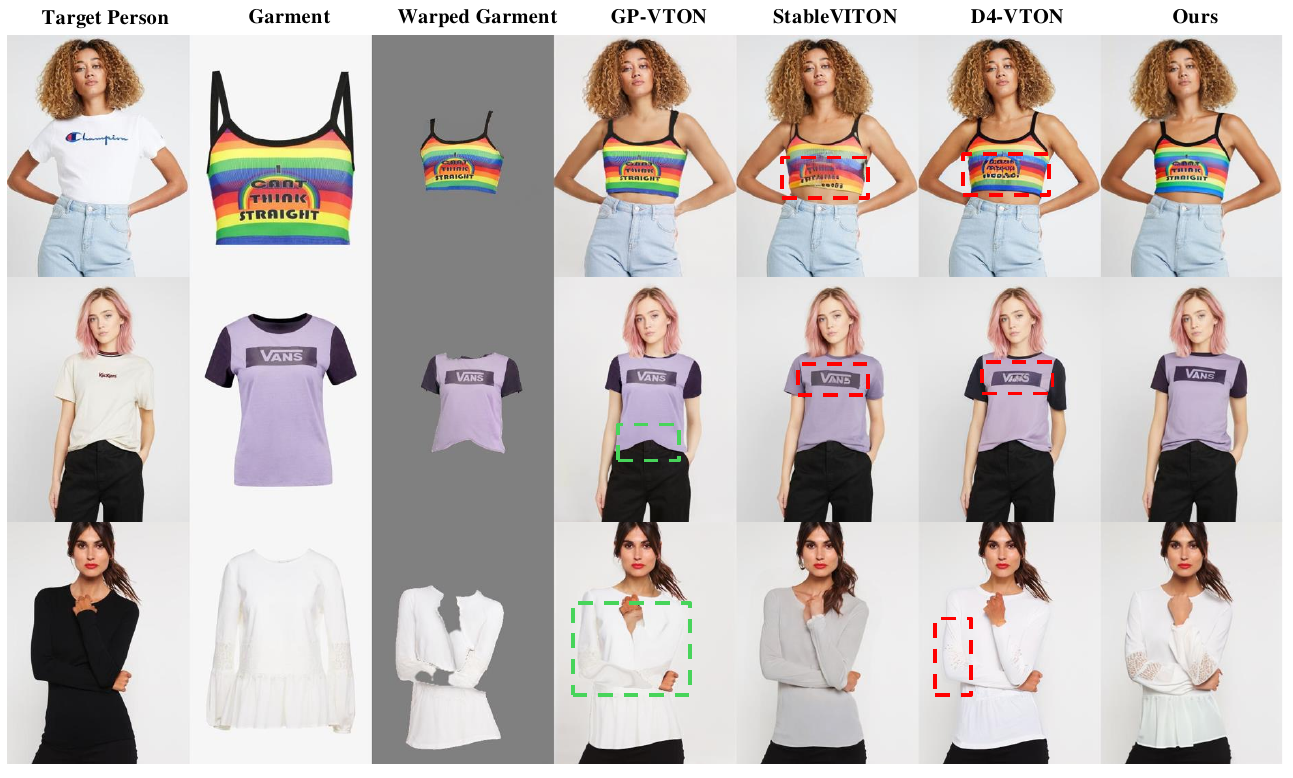}
 
\caption{Warp-based method (e.g. GP-VTON) is prone to severe visual artifacts and distortions (see the \green{green} dashed box) affected by the warp module.
Learning-based methods (e.g. StableVTON and $D^4$-VTON) are difficult to reconstruct complex detailed textures (see the \red{red} dashed box).
Our FIA-VTON designs a Flow Infused Attention module, utilizing the warping flow as an implicit guide to reconstruct complex detailed textures while maintaining the consistency of the garment.}
\label{fig:teasor}
\vspace{3mm}
}]

\begin{abstract}
\vspace{-8mm}

Image-based virtual try-on is challenging since the generated image should fit the garment to model images in various poses and keep the characteristics and details of the garment simultaneously. 
A popular research stream warps the garment image firstly to reduce the burden of the generation stage, which relies highly on the performance of the warping module. Other methods without explicit warping often lack \delong{sufficient} guidance to fit the garment to the model images. 
In this paper, we propose FIA-VTON, which leverages the implicit warp feature by adopting a Flow Infused Attention module on virtual try-on. The dense warp flow map is projected as indirect guidance attention to enhance the feature map warping in the generation process implicitly, which is less sensitive to the warping estimation accuracy than an explicit warp of the garment image. To further enhance implicit warp guidance, we \delong{incorporate} high-level spatial attention to complement the dense warp.
Experimental results on the VTON-HD and DressCode dataset significantly outperform state-of-the-art methods, demonstrating that FIA-VTON is effective and robust for virtual try-on. 

\begin{textblock*}{7cm}(1.5cm, 25.5cm)  
    \footnotesize{\noindent\rule{4cm}{0.4pt} \\
    $*$ Work done during internship at Shopee Inc. \\
    $\dagger$  Equal Contribution. \\
    $\ddagger$ Corresponding Author. \\
    }
\end{textblock*}

\end{abstract}

\vspace{-5mm}
\section{Introduction}

Virtual try-on technology generates images or videos of fitting a garment image on a model image, which is widely used in online shopping, special effects entertainment, AR, and other scenarios. Virtual try-on is challenging since the garment image needs to be warped properly to fit the pose of the model, keeping the garment consistent before and after try-on. At the same time, the generated image should be clear enough to maintain the garment details\cite{zhu2023tryondiffusion}. 

Along with the rapid progress of image generation techniques, the backbone of virtual try-on has evolved from GAN-based methods to diffusion models. \delong{However,} how to warp the garment properly remains under investigation. Previous methods can be roughly categorized into two research streams. \delong{Warping-based methods} warp the garment explicitly to fit the model's skeleton and then take the warped garment as input for the following generation process \cite{han2018viton,xie2023gp,wan2024improving}. In particular, many methods in the stream put the warped garment onto the cloth-agnostic model image, so the generation process becomes a typical inpainting problem \cite{gou2023taming,yang2024d}. However, these methods heavily rely on the performance of the warping module, which is highly challenging itself, as can be seen in the green dashed box of \Cref{fig:teasor}.
\delong{On the other hand,} \delong{Learning-based methods} take the garment image as a control condition\cite{zhu2023tryondiffusion,kim2024stableviton,chong2024catvton}. Both the warping and inpainting effects are produced simultaneously by a single-generation model. While these methods do not rely on a separate warping module, they lack direct guidance to the warping process. The try-on results will be degraded for model images and lack of complex detailed textures, which is depicted in the red dashed box of \Cref{fig:teasor}.

To overcome these limitations, we propose to leverage the implicit warp feature to guide the generation \delong{process}, which adopts a Flow Infused Attention module \delong{in} the diffusion model (FIA-VTON). 
Firstly, a dense flow map is predicted by a warp network from the garment image and the model pose. Then the dense flow is infused into the diffusion model by cross-attention. Unlike previous models which warp the garment explicitly by the flow map and input the warped garment to the generation model, flow attention takes the flow map itself to guide the implicit warp. In this way, the flow guidance is straight and clear. At the same time, the flow guidance is not as rigid as a warped garment, so the generation process can correct inaccurate flow estimations according to other sources of guidance such as the model pose. 

While the dense optical flow captures the deformation pattern for try-on, the static garment details are captured by \delong{the local garment feature} from a Garment Net, following Tryondiffusion\cite{zhu2023tryondiffusion}. To further capture high-level characteristics of the garment, we incorporate extra spatial attention into the flow attention. All the guidance, including the dense flow map, the local garment feature, and the high-level spatial feature are infused into the denoising UNet of the diffusion model by decoupled cross attention. As a result, the try-on image exhibits accurate deformations \delong{while preserving} both high-level characteristics and fine details.

We conduct extensive experiments on the VITON-HD~\cite{choi2021viton} and Dress-Code~\cite{morelli2022dress} dataset. The results demonstrate that FIA-VTON achieves state-of-the-art performance consistently. Qualitative results show that our model fitted the garment accurately to the model pose and maintained fine details as well.

The main contributions of this paper are as follows:
\begin{itemize}
    \item We propose to leverage Flow Infused Attention to guide the diffusion model, which acts as a straight and clear guide to the warping process of try-on.
    \item We infuse both the dense flow map, the local garment feature, and the high-level spatial feature by decoupled cross-attention, which captures the dynamic warp pattern, the garment details, and the high-level characteristics in a uniform way.
    \item We achieve state-of-the-art performance on widely used benchmarks, which validates the advantage of flow guidance over explicit guidance of warped garments and implicit warp without flow estimation.
\end{itemize}

\section{Related Work}

\noindent\textbf{\delong{Learning-based Virtual Try-on.}} 
\delong{Recently, diffusion models~\cite{ho2020denoising,dhariwal2021diffusion,ho2022classifier,gu2022vector} start to dominate in natural image generation due to its superior ability in generating high-fidelity realistic images compared to GAN-based models. Inspired by this, a series of learning-based virtual try-on models, which leverage diffusion framework to learn garment fitting and transformation, began to emerge.} TryOnDiffusion~\cite{zhu2023tryondiffusion} unifies two UNets to preserve garment details and warp the garment in a single network. 
LaDI-VTON~\cite{morelli2023ladi} introduces the textual inversion component that maps visual features of reference garment to CLIP token embedding space as a condition of diffusion model. 
DCI-VTON~\cite{gou2023taming} further use warping network to warp reference garment, which is fed into diffusion model as additional guidance.
StableVITON~\cite{kim2024stableviton} discarded independent warping and proposed a zero cross-attention block to learn semantic correlation between the clothes and human body.
Although promising results are attained, these learning-based VTON approaches still fail to completely retain every detail of the reference garment.
In contrast, our method introduces the Flow-infused Layer, which combines with implicit garment features to provide explicit guidance for fine-grained texture reconstruction.

\begin{figure*}[t]
    \vspace{-4mm}
    \centering
    \includegraphics[width=0.99\linewidth]{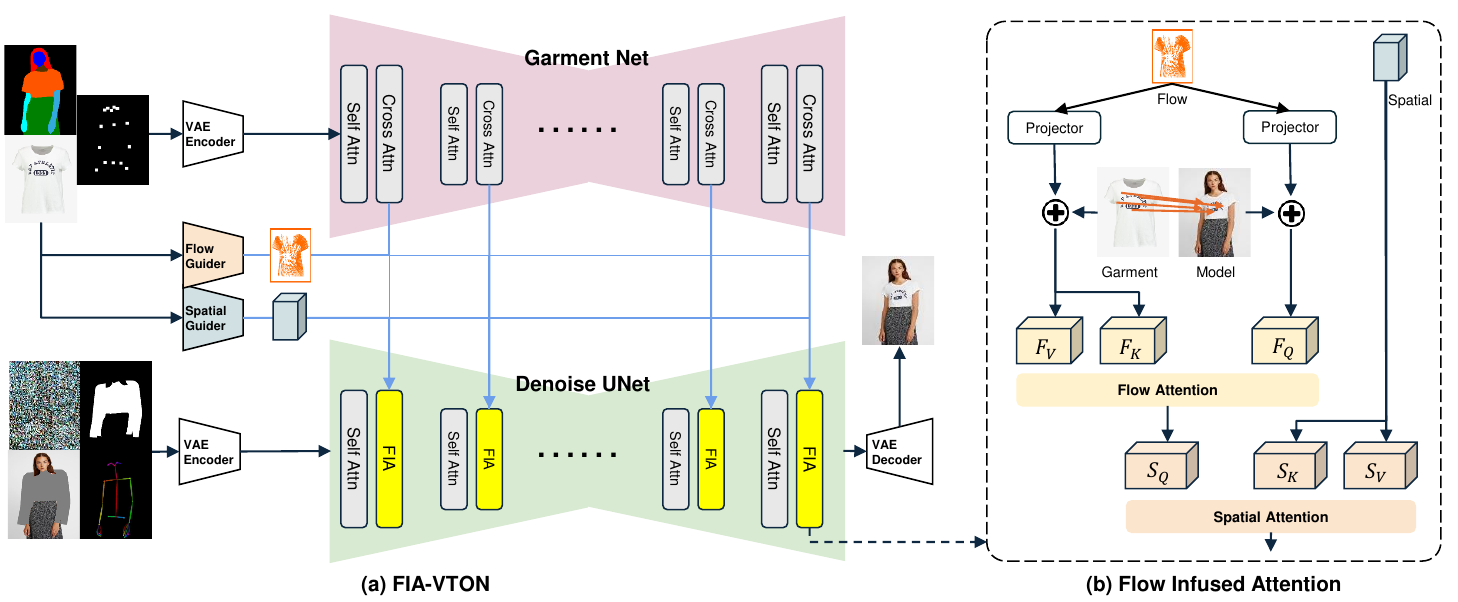}
    \caption{
    (a) Overview of FIA-VTON, illustrating the main components: a pair of VAE Encoder and Decoder, a Garment Net, and a Denoising UNet. The model takes a garment image, a target person image, and mask and pose based control images  as inputs. We adopt a Flow Guider to generate dense flow map, and a Spatial Guider to extract global garment characteristics. These features are processed in a Flow Infused Attention module, interactions with Garment Net and Denoising UNet to generate the final try-on output. (b) Illustration of the Flow Infused Attention module. The flow is projected into garment and model feature space, and then fused using a cross-attention mechanism to produce flow-guided features. Then the high-level spatial feature is further integrated by decoupled cross-attention, which captures the garment consistency, details, and high-level characteristics uniformly.
    }
    \label{fig:method_overview}
    \vspace{-2mm}
\end{figure*}

\noindent\textbf{Wrap Guidance.}
Most virtual try-on methods have a pre-warping stage, which warps flat garments to align with the corresponding positions on the human body. DCI-VTON \cite{gou2023taming} and $D^4$-VTON \cite{yang2024d} paste the warped garments onto the cloth-agnostic model image, adds noise, and then inputs into a denoising diffusion model. GP-VTON \cite{xie2023gp} concatenates the warped garment with cloth agnostic model image and other guide images spatially, and inputs them into a UNet generator. GarDiff\cite{wan2024improving} concatenates the warped garment with the noised image and feeds them into the self-attention layer of a denoising diffusion model. In contrast, our model injects the dense optical flow instead of the warped garment into the diffusion model through cross attention, which is softer and thus less sensitive to the warp estimation precision.
GarDiff also combines the Embeddings of warped Garment and raw garment image. ALDM\cite{gao2024aldm} align the features of the raw garment and the warped garment by an Adaptive Alignment Module, and then input the aligned feature into the diffusion model by cross attention. These combined features cannot reflect the warping pattern directly. In comparison, our model uses the flow map instead, which can give clear and direct guidance to warping.

\noindent\textbf{Flow Estimation.}
Early try-on methods \cite{wang2018toward,li2021toward} employ Thin Plate Spline (TPS) to warp flat garments to align with the corresponding positions on the human body. To better handle complex deformations, appearance flow-based methods \cite{han2019clothflow,he2022style,ge2021parser,yang2024d}  have been introduced to predict appearance flow to warp the flat garment to achieve pixel-level alignment with the target person’s body. $D^4$-VTON \cite{yang2024d} proposed a Dynamic Semantics Disentangling Module to predict flows separately for different semantic layers. Our paper follows $D^4$-VTON to extract dense warp flows. 

\vspace{-3mm}
\section{Our method}
In this work, we propose a Flow Infused Attention module for diffusion-based virtual try-on (FIA-VTON). Figure \ref{fig:method_overview} demonstrates an overview of the FIA-VTON model. FIA-VTON is built on an inpainting version of Stable Diffusion (SD)~\cite{rombach2022high}. SD first encodes the noised image and the masks by a VAE encoder, then generates a latent image by a Denoising Unet, and decodes into a real image by a VAE decoder. In FIA-VTON, all the cross-attention modules of the Denoising UNet are replaced by the Flow Infused Attention (FIA) modules. A FIA module fuses the dense warp flow, the local garment feature, and the coarse spatial feature into the model feature from the upstream UNet layers.  Here the local garment feature is extracted by a Garment Net, and we adopt the typical structure of a VAE Encoder and a UNet. The dense flow is estimated by a Flow Guider as described in Section \ref{sec:flow_guider}. The coarse spatial feature is provided by a Spatial Guider, detailed in Section \ref{sec:spatial_guider}.

\subsection{FIA module}
\label{sec:Flow_infused_layer}

Previous \delong{learning-based VTON approaches} struggle to preserve the details of garments, such as high-frequency textures in complex patterns. In addition, warp-based methods are prone to serve visual artifacts and distortions affected by the garment warp module. To solve these problems, we leverage the appearance flow and spatial characteristics, combining with the local garment feature to implicitly guide the diffusion model, in order to integrate the two previously mentioned approaches.

The FIA module takes four inputs. The first one is the dense warp flow $F \in \mathbb{R}^{ h \times w \times 2}$ between the original garment image and the garment in the target try-on image, where $h$ and $w$ are the height and width of the garment image, respectively. The second input is the garment feature $G$ extracted by the Garment Net. The third input is the model feature $P$ from the upstream layer of the Denoising Unet. The last input is the spatial embeddings $S$ generated by the Spatial Guider. The detail architecture is depicted in Figure~ \ref{fig:method_overview}.

Firstly, both the model feature $P$ and garment feature $G$ are combined with the projection of the warp flow $F$ by element-wise summation, as follows:
{\small
\begin{equation}
\begin{aligned}
    P_f &= P + F_P, \\
    G_f &= G + F_G, 
\end{aligned}
\end{equation}
}
where $F_P$ is the flow feature produced by an MLP projector that aligns the optical flow $F$ with the model feature $
P$. Similarly, $F_G$ is the flow features aligned with the garment feature $G$. Then, the combined features $P_f$ and $G_f$ are infused into the Denoising Unet by joint cross attention.
{\small
\begin{equation}
    P_e = \text{softmax} \left( \frac{F_Q(P_f) \cdot (F_K(G_f))^{\top}}{\sqrt{d_f}} \right) \cdot F_V(G_f),
\end{equation}
}
where the matrices $F_Q$, $F_K$, and $F_V$ are standard cross-attention weights for query, key, and value, respectively. The denominator $\sqrt{d_f}$ is the standard scale factor. The output is a model feature $P_e$ modulated by the dense optical flow and the garment feature. The flow $F$ acts as an offset of the intermediate features of the Denoising UNet that turns \( P \rightarrow P_e \), which brings spatial deformation guidance for implicit warping and inpainting.

Secondly, the modulated feature $P_e$ is further modulated by the spatial embeddings $S$. Specifically, given the garment image $\mathbf{I}_g$, the spatial embeddings $S$ is calculated as:
\begin{equation}
\begin{aligned}
    {S} &= \mathbf{S}_v(I_g),
\end{aligned}
\end{equation}
where $\mathbf{S}_v$ is the Spatial Guider encoder. Different from the garment feature $G$, the embedding $S$ focuses on high-level spatial information at different granularity. Formally, the spatial embedding $S$ is integrated with the enhanced model feature $P_e$ by a cross-attention layer, as follows:
\begin{equation}
    P_o = \text{softmax} \left( \frac{S_Q(P_e) \cdot (S_K(S))^{\top}}{\sqrt{d_s}} \right) \cdot S_V(S),
\end{equation}
where the matrices $S_Q$, $S_K$, and $S_V$ are standard cross-attention weights for query, key, and value respectively. The denominator $\sqrt{d_s}$ is the standard scale factor. 

We replace all cross-attention layers of the Denoising Unet with the FIA modules. In doing so, it supplements the latent space of the diffusion model with positional information of texture details, ensuring that all critical details are more accurately reconstructed.

\noindent\textbf{Discussion.}
There several possible ways to combine the warp flow and the garment feature and the model feature, such as pixel-wise multiplication, concatenation, and so on. In practice the warp flow often concentrates in local areas with large motion, which is sparse. So we choose feature summation for numeric stability and simplicity.
This implicit fusion of flow and diffusion features not only corrects potential deformation errors but also ensures that the final try-on results maintain both accurate alignment and detailed texture preservation, demonstrating superior experimental performance.
Detailed ablation study are given in \Cref{sec:ablation}.

\subsection{Flow Guider}
\label{sec:flow_guider}
The Flow Guider estimates the dense warp flow $F \in \mathbb{R}^{ h \times w \times 2}$ from the original garment image to the target garment in try-on image. In this paper we adopt the Dynamic Semantics Disentangling
Module (DSDM) from $D^4$-VTON \cite{yang2024d}. DSDM takes as input the garment image and the condition triplet $T$ including the human pose, the DensePose pose, and the preserve region mask. Two feature pyramid networks (FPN) are used to extract multi-scale features, and DSDM  generates local flows for different semantic regions from coarse to fine. In practice we utilize the final flow as an input to FIA, and keep the DSDM freezed during training FIA-VTON.



\subsection{Spatial Guider}
\label{sec:spatial_guider}
The Spatial Guider aims to extract high-level spatial features from the garment image. In comparison, the garment feature $G$ extracted by a UNet focus on low-level image features. These two features are complementary, so they can fully excavate the information of the garment.

We adopt FashionCLIP as the Spatial Guider in practice. FashionCLIP is fine-tuned on millions of image-text pairs of fashion clothes, so it achieves better understanding on the patterns of the clothes textures and complex texts than vanilla CLIP. Since the Spatial Guider needs to extract high-level appearance information while keeping spatial information, we select the highest level feature map without visual projection.


\subsection{FIA-VTON}
The FIA-VTON model can be considered as an inpainting version of Stable Diffusion with control signals infused by the FIA modules in Section \ref{sec:Flow_infused_layer}. Specifically, given a model image $\mathbf{I}_p \in \mathbb{R}^{H \times W \times 3}$ and garment image $\mathbf{I}_g \in \mathbb{R}^{H \times W \times 3}$, our goal is to synthesize them into a realistic image $\mathbf{\hat{I}} \in \mathbb{R}^{H \times W \times 3}$, which has the same person attributes as in $\mathbf{I}_p$ while retaining the garment from $\mathbf{I}_g$.

The input to the Denoising Unet is a 12-channel tensor composed of 4 channels of latent noise $\epsilon$, 4-channel masked person latent $x_m$, 1-channel mask $\mathbf{m}$, and 3-channel of the skeleton $\mathbf{s}$. 
The $x_m \in \mathbb{R}^{h \times w \times 3}$ is generated from $\mathbf{I}_p$ using OpenPose~\cite{cao2017realtime} and HumanParsing~\cite{li2020self}, and go through a VAE encoder to transform it into the latent space, where $h = \frac{H}{8}$ and $w = \frac{W}{8}$. On the other side, we feed the garment images $\mathbf{I}_g$ to the Garment Net to obtain the local garment feature in a single time step. The Garment Net is a network similar to a denoising UNet. During training, \delong{we} initialize the Garment Net and the denoising UNet by the same weights to ensure the consistency between the garment's local texture features and the denoised image features.

Along with the aforementioned auxiliary conditioning input, the Garment Net and the denoising UNet are
jointly trained by minimizing the following loss function:
\begin{equation}
\delong{\mathcal{L}_{DM}} = \mathbb{E}_{\mathbf{x}_t, \mathbf{x}_g, F,S, \epsilon, t} \left( \|\epsilon_\theta(\mathbf{x}_t, \mathbf{x}_g, F,S, ) - \epsilon\|_2^2 \right),
\end{equation}
where $\mathbf{x}_g$ is the latent feature of the garment image from the VAE encoder, $\mathbf{x}_t$ is the noised latent feature of the try-on image at time step $t$ with noise $\epsilon$, $\theta$ is the parameters of denoising UNet model and the garment model, $S$ is the high-level feature of Fashion-CLIP, and $F$ is the predicted flow of the warping module.

\section{Experiments}

In this section, we present an evaluation of our proposed FIA-VTON
method on two widely studied datasets ~\cite{choi2021viton, morelli2022dress}, and compare it to various state-of-the-art approaches, including Warp-based (VITON-HD~\cite{choi2021viton}, HR-VTON~\cite{lee2022high}, GP-VTON~\cite{xie2023gp}, DCI-VTON~\cite{gou2023taming}, $D^4$-VTON~\cite{yang2024d}), and Learning-based methods (LADI-VTON~\cite{morelli2023ladi}, StableVITON~\cite{kim2024stableviton}, IDM-VTON~\cite{choi2024improving}, GarDiff~\cite{wan2024improving}). Additionally, we
investigate the effect of implicit attention modules in FIA to demonstrate the robustness of our method.

\begin{figure*}[t]
    \centering
    \includegraphics[width=0.95\linewidth]{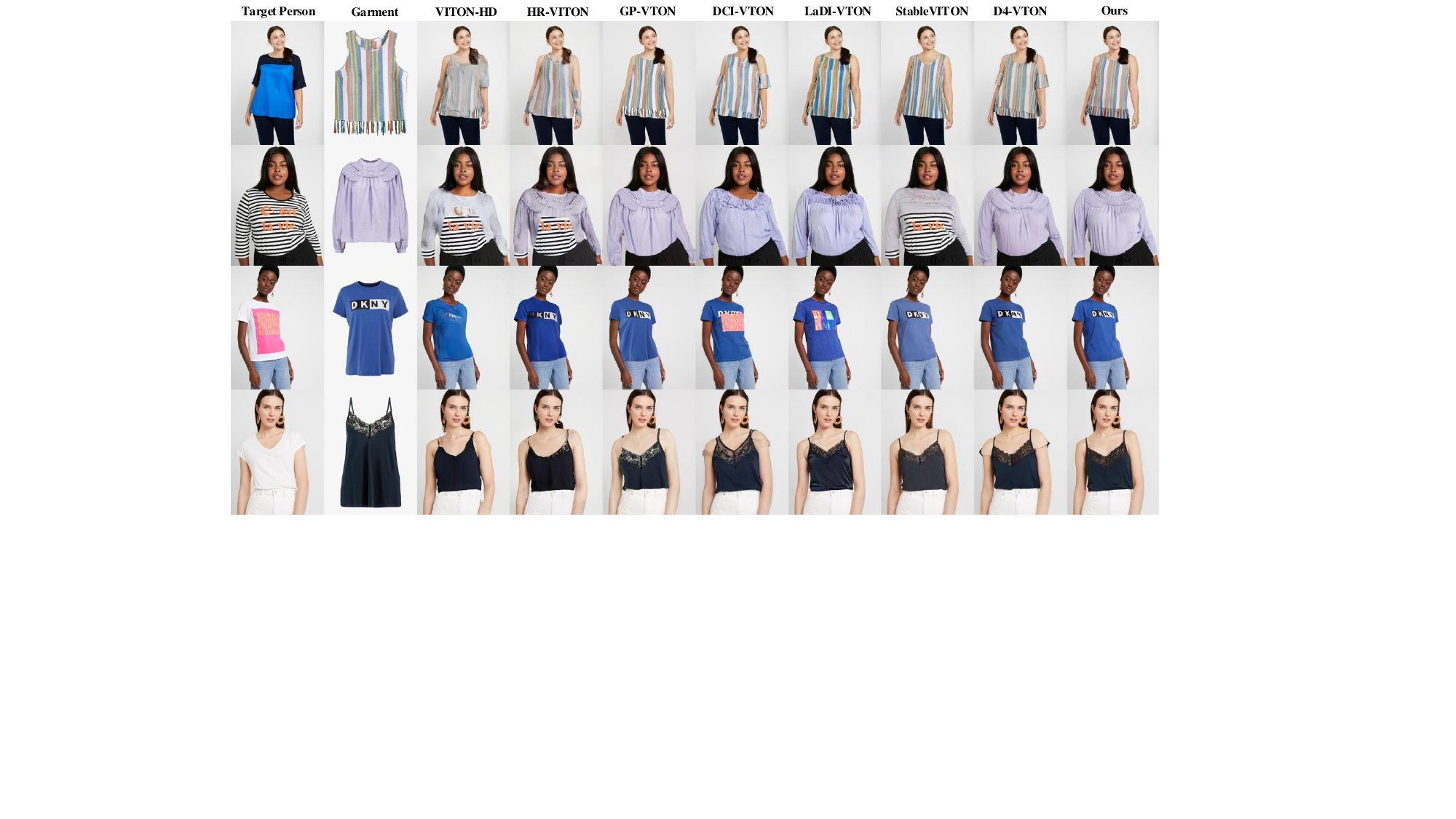}
    \vspace{-4mm}
    \caption{Qualitative comparison on VITON-HD dataset~\cite{choi2021viton}. Examples generated by VITON-HD, HR-VTON, GP-VTON, LaDI-VTON, DCI-VTON, StableVITON, $D^4$-VTON and our model. Zoom in for a better view.}
    \label{fig:methods_comparison}
\end{figure*}

\subsection{Experiments Setup}
\noindent\textbf{Datasets.} 
There are two commonly used benchmarks for evaluating high-resolution virtual try-on: VITON-HD~\cite{choi2021viton} and Dress-Code~\cite{morelli2022dress}. These two datasets exhibit significant differences in the poses and styles of the models. VITON-HD consists of 13,679 image pairs of front-view upper-body model and upper-body garment images, which are further split into 11,647 training pairs and 2,032 testing. DressCode contains 53792 image pairs of front-view full-body person and in-store garment images, which are composed of three subsets with different category pairs, that is upper, lower, and dresses. It has 48,392 training set and 5,400 testing set. We follow the experimental setup of previous methods \cite{morelli2023ladi, kim2024stableviton} and conduct experiments under the image resolution of 512 × 384. 

\noindent\textbf{Evaluation metrics.}
We conduct evaluations with two testing settings: paired and unpaired. In the paired setting, a garment image is used to reconstruct the image of the person originally wearing it, while in the unpaired setting, the garment worn by the person is replaced with a different one.
For both evaluation settings, we adopt a widely used metric: Frechet Inception Distance (FID)~\cite{parmar2022aliased}.
Moreover, in the paired setting where ground truth is available for comparison, we employ Structural Similarity (SSIM)~\cite{binkowski2018demystifying}, Peak Signal-to-Noise Ratio (PSNR) and Learned Perceptual Image Patch Similarity (LPIPS)~\cite{zhang2018unreasonable} to thoroughly evaluate the performance of the try-on results. 
Note that we find different downsampling methods have a significant impact on both paired and unpaired FID scores. Unless otherwise specified, our experiments use the bilinear interpolation setup from GP-VTON and $D^4$-VTON as the default.
Additionally, we present qualitative results, providing a comparison with previous methods to further demonstrate the effectiveness of our approach.

\noindent\textbf{Implementation Details.}
We train our models based on the inpainting version's pre-trained StableDiffusion v2.1\cite{huggingface_stable_diffusion}. 
For a fair comparison, we train two separate models for the VITON-HD and DressCode datasets, evaluating them on their respective test sets as done in previous methods. 
All models are trained with identical hyperparameters: the AdamW \cite{loshchilov2017fixing} optimizer is used with a batch size of 64 and a constant learning rate of $1e-5$ for 100,000 steps at a resolution of $512 \times 384$. 
All experiments are conducted on 4 NVIDIA H100 GPUs, with each model requiring approximately 80 hours of training.

\begin{table}[]
    \centering
    \caption{Quantitative results on VITON-HD dataset. We compare with Warp-based and Learning-based virtual try-on methods. FID$_p$/FID$_u$ stands for the FID score in the paired/unpaired setting. We follow the data processing approach of GP-VTON and $D^4$-VTON, using bilinear interpolation for downsampling. An asterisk (\textbf{*}) indicates cases where cubic downsampling was applied. 
    \textbf{Bold} denotes the best score for each metric and \underline{Underlining} indicates the second best result. }
    \label{Tab:vitonhd_results}
    \setlength{\tabcolsep}{0.9mm}{
    \resizebox{0.46\textwidth}{!}{
        \begin{tabular}{lccccc}
        \toprule
        \multicolumn{1}{l|}{Method}         & SSIM$\uparrow$ & \multicolumn{1}{l}{PSNR$\uparrow$} & LPIPS$\downarrow$ & $\mathbf{FID_p}\downarrow$ & $\mathbf{FID_u}\downarrow$ \\ \midrule
        \multicolumn{6}{c}{\cellcolor{lightgray}Warp-based methods}                                                                                           \\
        \multicolumn{1}{l|}{VITON-HD}       & 0.862          & -                                   & 0.117             & -                  & 12.12              \\
        \multicolumn{1}{l|}{HR-VTON}        & 0.878          & 21.61                              & 0.105             & 11.383             & 11.27              \\
        \multicolumn{1}{l|}{GP-VTON}       & 0.884          & 23.41                              & 0.081             & 6.031              & 9.072  
        \\
        \multicolumn{1}{l|}{DCI-VTON}       & 0.880          & 24.01                              & 0.080             & 5.521              & 8.754              \\
        \multicolumn{1}{l|}{$D^4$-VTON}       & 0.892          & 24.71                              & 0.065             & {\uline{4.845}}        & 8.53               
        \\ \midrule
        \multicolumn{6}{c}{\cellcolor{lightgray}Learning-based methods}                                                                                   \\ 
        \multicolumn{1}{l|}{LaDI-VTON}      & 0.864          & 22.49                              & 0.096             & 6.602              & 9.48               \\
        \multicolumn{1}{l|}{StableVITON}    & 0.888          & -                                  & 0.073             & -                  & 8.233              \\
        \multicolumn{1}{l|}{IDM-VTON}       & 0.870          & -                                  & 0.102             & -                  & 8.64               \\
        \multicolumn{1}{l|}{GarDiff}        & {\uline{0.912}}    & -                                  & \textbf{0.036}    & 6.021              & {\uline{7.89}}         \\ \midrule
        \multicolumn{1}{l|}{\textbf{Ours*}}  & 0.906          & {\uline{27.355}}                       & {\uline{0.047}}       & 5.124              & \textbf{7.869}     \\
        \multicolumn{1}{l|}{\textbf{Ours}} & \textbf{0.913} & \textbf{27.827}                    & {\uline{0.047}}       & \textbf{4.686}     & 7.914              \\ \bottomrule
        \end{tabular}
        }
        }
        \vspace{-7mm}
\end{table}

\begin{table*}[]
\centering
\caption{Quantitative results on DressCode datasets. FID$_p$/FID$_u$ stands for the FID score in the paired/unpaired setting. \textbf{Bold} denotes the best score for each metric and \underline{Underlining} indicates the second best result.}
\vspace{-3mm}
\label{Tab:dresscode_results}
    \resizebox{0.9\textwidth}{!}{
        \begin{tabular}{lcccccccccccc}
        \toprule
        \multicolumn{1}{c|}{}                         & \multicolumn{4}{c|}{DressCode-Dresses}                                                      & \multicolumn{4}{c|}{DressCode-Lower}                                                        & \multicolumn{4}{c}{DressCode-Upper}                                        \\ \cmidrule{2-13} 
        \multicolumn{1}{c|}{\multirow{-2}{*}{Method}} & SSIM$\uparrow$ & LPIPS$\downarrow$ & $\mathbf{FID_u}\downarrow$ & \multicolumn{1}{c|}{$\mathbf{FID_p}\downarrow$} & SSIM$\uparrow$ & LPIPS$\downarrow$ & $\mathbf{FID_u}\downarrow$ & \multicolumn{1}{c|}{$\mathbf{FID_p}\downarrow$} & SSIM$\uparrow$ & LPIPS$\downarrow$ & $\mathbf{FID_u}\downarrow$ & $\mathbf{FID_p}\downarrow$     \\ \midrule
        \multicolumn{13}{c}{\cellcolor{lightgray}Warp-based methods}                                                                                               \\
        \multicolumn{1}{l|}{HR-VITON}                 & 0.865          & 0.113             & 18.81           & \multicolumn{1}{c|}{16.82}           & 0.937          & 0.045             & 16.39           & \multicolumn{1}{c|}{11.41}           & 0.916          & 0.071             & 16.82           & 15.37               \\
        \multicolumn{1}{l|}{GP-VTON}                  & 0.881          & 0.073             & 12.64           & \multicolumn{1}{c|}{7.44}            & 0.941          & 0.042             & 16.07           & \multicolumn{1}{c|}{7.73}            & 0.947          & 0.036             & 11.98           & 7.38     
        \\
        \multicolumn{1}{l|}{DCI-VTON}                 & 0.887          & 0.070             & 12.35           & \multicolumn{1}{c|}{8.48}            & 0.939          & 0.045             & 15.45           & \multicolumn{1}{c|}{7.97}            & 0.942          & 0.041             & 11.64           & 7.47               \\
        \multicolumn{1}{l|}{$D^4$-VTON}                  & 0.890          & {\uline{0.061}}       & 12.29           & \multicolumn{1}{c|}{{\uline{8.28}}}      & {\uline{0.946}}    & {\uline{0.033}}       & 14.86           & \multicolumn{1}{c|}{{\uline{6.67}}}      & 0.942          & 0.041             & {\uline{11.00}}     & {\uline{6.55}}       
        \\ \midrule
        \multicolumn{13}{c}{\cellcolor{lightgray}Learning-based methods}                                                                                                                                                                                                                                                 \\
        \multicolumn{1}{l|}{LaDI-VTON}                & 0.868          & 0.089             & 13.40           & \multicolumn{1}{c|}{-}               & 0.922          & 0.051             & 14.80           & \multicolumn{1}{c|}{-}               & 0.928          & 0.049             & 13.26           & -                   \\
        
        \multicolumn{1}{l|}{StableVITON}              & -              & 0.068             & 12.25           & \multicolumn{1}{c|}{-}               & -              & 0.035             & 12.34           & \multicolumn{1}{c|}{-}               & 0.937          & 0.039             & \textbf{9.94}   & -                   \\
        
        \multicolumn{1}{l|}{GarDiff}                  & {\uline{0.891}}    & 0.065             & {\uline{12.05}}     & \multicolumn{1}{c|}{8.77}            & 0.939          & 0.035             & {\uline{12.29}}     & \multicolumn{1}{c|}{8.01}            & \textbf{0.952} & {\uline{0.030}}       & 11.32           & 8.69        
        \\ \midrule
        \multicolumn{1}{l|}{Ours}                     & \textbf{0.901} & \textbf{0.049}    & \textbf{10.45}  & \multicolumn{1}{c|}{\textbf{6.85}}   & \textbf{0.952} & \textbf{0.027}    & \textbf{10.58}  & \multicolumn{1}{c|}{\textbf{6.02}}   & \textbf{0.952}    & \textbf{0.027}    & {\uline{10.76}}  & \textbf{6.22} \\ \bottomrule
        \end{tabular}
        }
    \vspace{-3mm}
\end{table*}

\subsection{Comparison with the state-of-the-arts}
As presented in Table \Cref{Tab:vitonhd_results} and \Cref{Tab:dresscode_results}, our proposed FIA-VTON demonstrates superior performance compared to previous state-of-the-art methods, with a significant margin across all two datasets and in various scenarios.

\noindent\textbf{VITON-HD.} We compare the state-of-the-art Warp-based and Learning-based virtual try-on methods on the VITON-HD dataset. 
The quantitative comparison is presented in \Cref{Tab:vitonhd_results}. Our method achieves an SSIM/PSNR score of 0.913/\delong{27.827}, significantly outperforming other methods and demonstrating superior consistency between the generated garment and the model. 
Moreover, our method achieves outstanding performance in the FID$_p$ metric, with a best score of 7.869, indicating a reduced perceptual distance between the generated images and real ones.
DCI-VTON and $D^4$-VTON explicitly combine the warping module and diffusion model, which is the most relative to our method. However, they are strongly dependent on the performance of the warping results. This dependency leads to incorrect garment regions when dealing with substantial variations or significant transformations in garment style. DCI-VTON and $D^4$-VTON perform worse than our method by 0.885 and 0.661 on the $\mathbf{FID_u}$ metric, respectively. 
Also when compared to Learning-based methods, our method is still able to achieve more accurate fine-grained garment deformation, enabling better recovery of garment details. That our SSIM score exceeds LdDI-VTON and \delong{StableVITON} by 0.049 and 0.025. 

Furthermore, we depict the qualitative comparison with other methods on the VITON-HD in \Cref{fig:methods_comparison}. As can be seen, Warp-based methods like HR-VITON and GP-VTON often struggle to generate realistic human body parts, such as abdomens or arms (row 1), resulting in noticeable ``copy and pasts'' artifacts.
The underlying reason is that Warp-based approaches rely heavily on the effectiveness of the warp model and cannot handle overly vigorous deformations.
Among Learning-based methods, LADI-VTON, and StableVITON fail to capture accurate garment texture (rows 2,3) due to a lack of guidance of texture information, leading to texture details that differ significantly from the target garment.
DCI-VTON and $D^4$-VTON incorporate warping module and diffusion model to generate natural dressing results. However, they still use explicit warping as a priori information and are dependent on the performance of the warping module.
When confronted with drastic deformation of clothing styles, wrong clothing artifacts are usually generated (row 4).
In contrast, our FIA-VTON applies the flow generated by the warping module as an implicit guide, skillfully integrating it into the attention layer. It effectively prompts the diffusion model to reconstruct texture details while adeptly handling significant garment deformations.

\begin{figure}[t]
    \centering
    \includegraphics[width=0.95\linewidth]{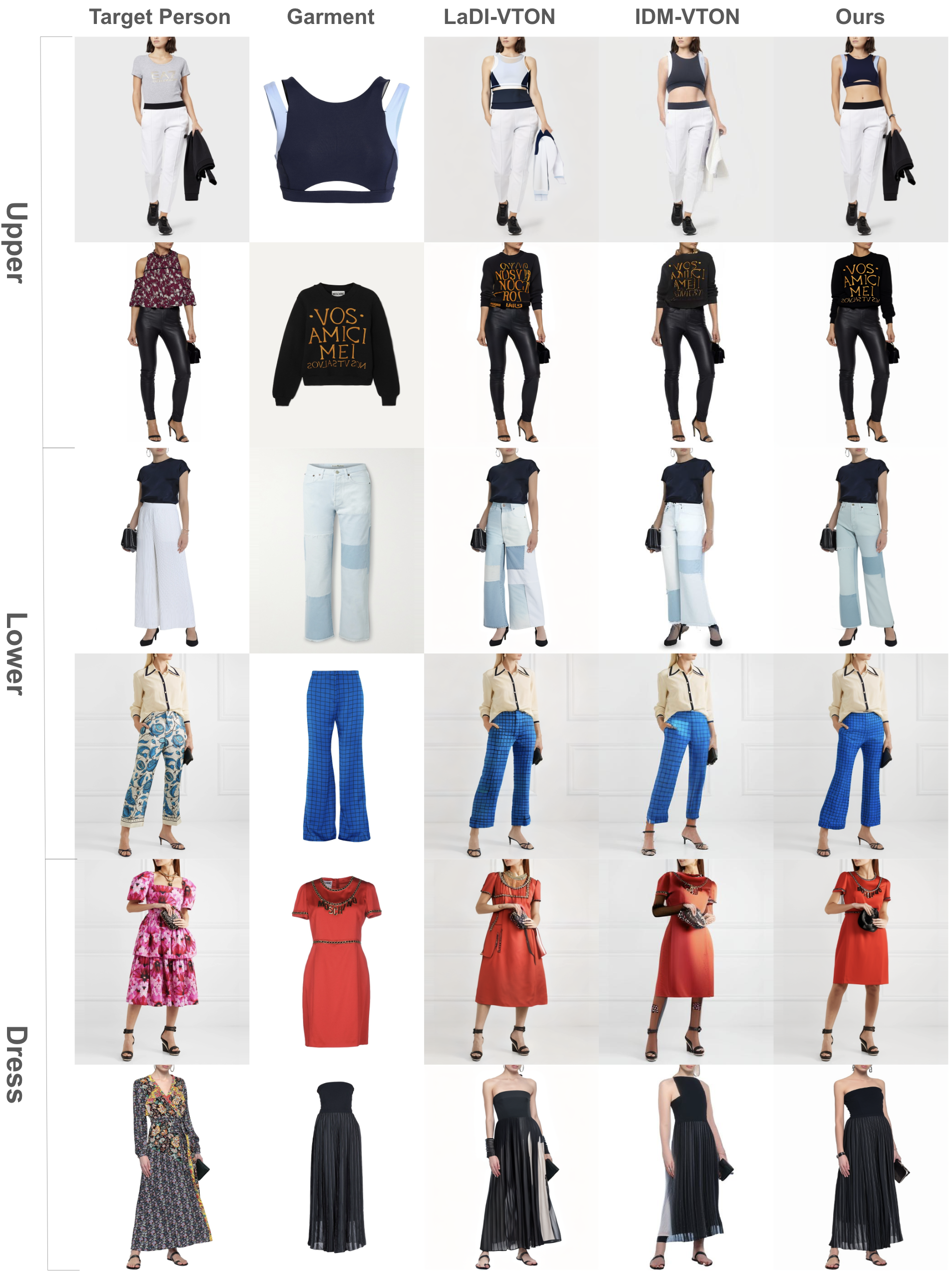}
    \vspace{-3mm}
    \caption{
        Qualitative comparison on the DressCode~\cite{morelli2022dress} dataset. FIA-VTON demonstrates a distinct advantage in handling complex textures and Drastic deformation. Please zoom in for more details.
    }
    \label{fig:dresscode}
    \vspace{-5mm}
\end{figure}

\noindent\textbf{DressCode.}
\Cref{Tab:dresscode_results} summarizes the performance comparisons on the DressCode dataset. FIA-VTON surpasses other competing methods across all evaluation metrics on ``DressCode-Dresses'' and ``DressCode-Lower'' test sets, emphasizing its ability to generate natural fitting results and preserve fine-grained garment textures. 
\Cref{fig:dresscode} illustrates the comparison for different garment types (including upper, lower, and dress) on full-body person images from the DressCode dataset. For uppers, akin to the results in ~\Cref{fig:dresscode}, our approach can generate results more consistent with the garment textures, devoid of artifacts.
Regarding lowers and dresses, our method can accurately recognize the type and texture of the garments \delong{(row 3-5)}, and perform better in rendering semi-transparent materials \delong{(row 6)}.

\begin{table}[]
\caption{
Ablation study on each module of FIA-Diff on VITON-HD. 
    The table compares our model's performance with and without the Flow Guider or Spatial Guider. Removing either module leads to degradation in all metrics. 
}
\resizebox{0.48\textwidth}{!}{
    \begin{tabular}{l|ccccc}
    \hline
    Model                   & $\mathbf{FID_u}\downarrow$ & $\mathbf{FID_p}\downarrow$ & SSIM$\uparrow$ & PSNR$\uparrow$  & LPIPS$\downarrow$ \\ \hline
    w/o Flow Guider    & 8.530              & 5.438              & 0.902          & 26.545          & 0.055             \\
    w/o Spatial Guider & 8.645              & 6.029              & 0.888          & 24.330          & 0.076             \\
    w/ openclip & 8.032              & 4.693              & \textbf{0.914}          & \textbf{27.925}          & \textbf{0.047}             
    \\ \midrule
    Concat Input             & 8.204              & 4.950              & 0.912          & 27.604          & 0.047             \\
    Cross-Attention          & 8.31               & 5.28               & 0.904          & 26.945          & 0.051             
    \\ \midrule
    Ours                    & \textbf{7.914}     & \textbf{4.686}     & 0.913 & 27.827 & \textbf{0.047}    \\ \hline
    \end{tabular}
\label{tab:ablation_study}
}
    \vspace{-4mm}
\end{table}

\subsection{Ablation Study}
\label{sec:ablation}

\begin{figure}
    \centering
    \includegraphics[width=0.98\linewidth]{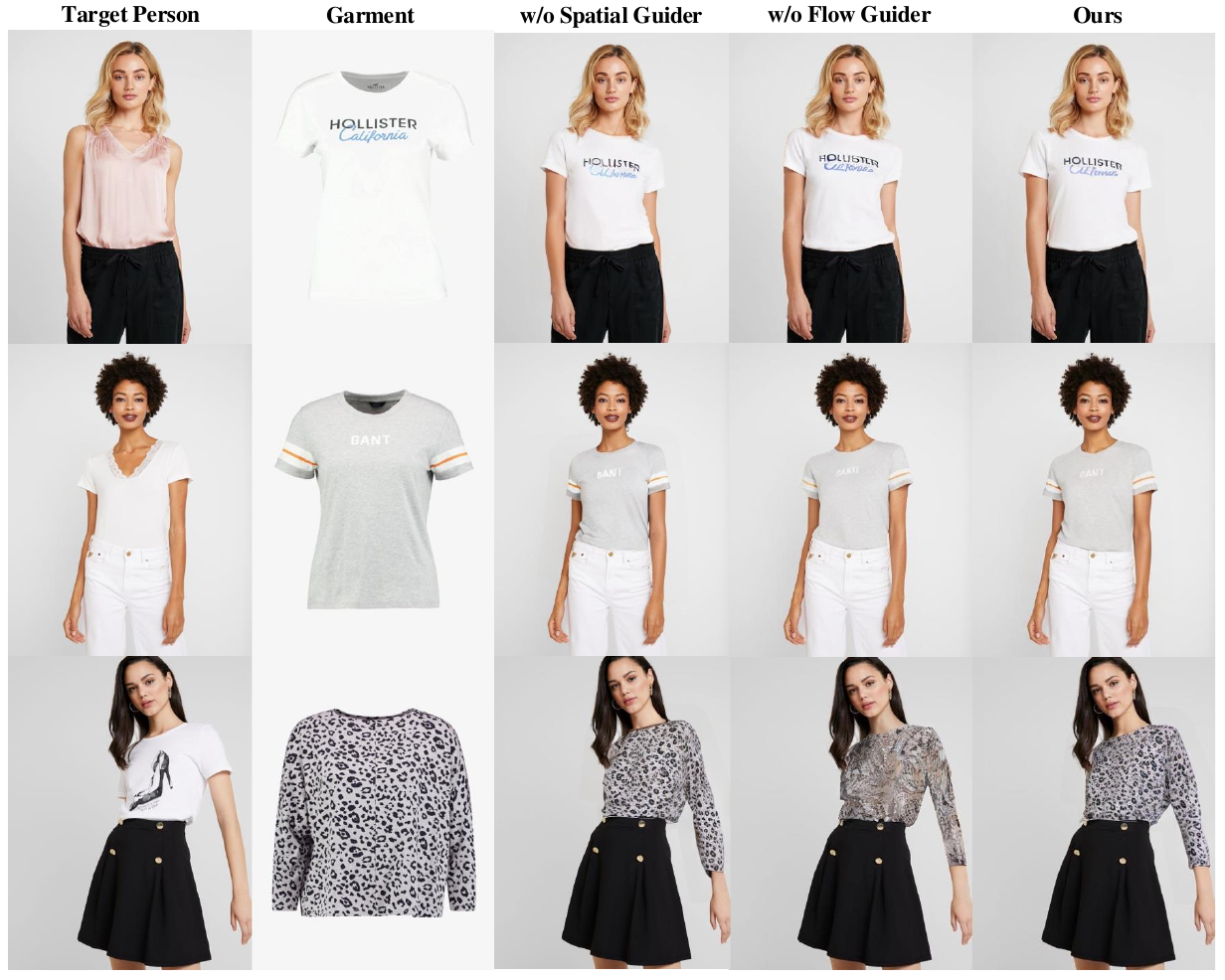}
    \vspace{-3mm}
    \caption{
    Ablation study on our FIA-VTON.
    }
    \label{fig:ablation_study}
    \vspace{-4mm}
\end{figure}

\Cref{tab:ablation_study} and \Cref{fig:ablation_study} present the ablation study results on the FIA-VTON model, analyzing the contributions of two key components: Flow Guider and Spatial Guider. The ablation study includes three settings: (1) removing the Flow Encoder, (2) removing the Spatil Guider, (3) replacing Fashion-CLIP with OpenCLIP, and (4) the full model.

\noindent\textbf{Effectiveness of the Flow Guider module.}
As shown in \Cref{tab:ablation_study},
compared to the model without the Flow guider module, the full model (Ours) achieves better performance on all metrics, particularly on FID$_p$.
The FID$_p$ is reduced from 5.438 to 4.686. 
In the 1st row of \Cref{fig:ablation_study} we observe that the lack of the Flow Guider makes the textual information in the generated results illegible.
This highlights the significant role of the Flow Guider in capturing texture semantics to accurately reconstruct garment details. 

\noindent\textbf{Effectiveness of the Spatial-Guider module.}
From ``w/o Spatial Guider'' of \Cref{tab:ablation_study}, we can observe that removing the Spatial Guider leads to increased $\mathbf{FID_u}$ (from 7.914 to 8.645) and decreased SSIM (from 0.913 to 0.888), indicating degraded image quality and alignment. Qualitative results in \Cref{fig:ablation_study} further confirm that without the Spatial Guider, generated garments exhibit misalignment and less coherent blending with the target person, particularly for garments with complex textures or patterns.

\begin{figure}
    \centering
    \includegraphics[width=0.96\linewidth]{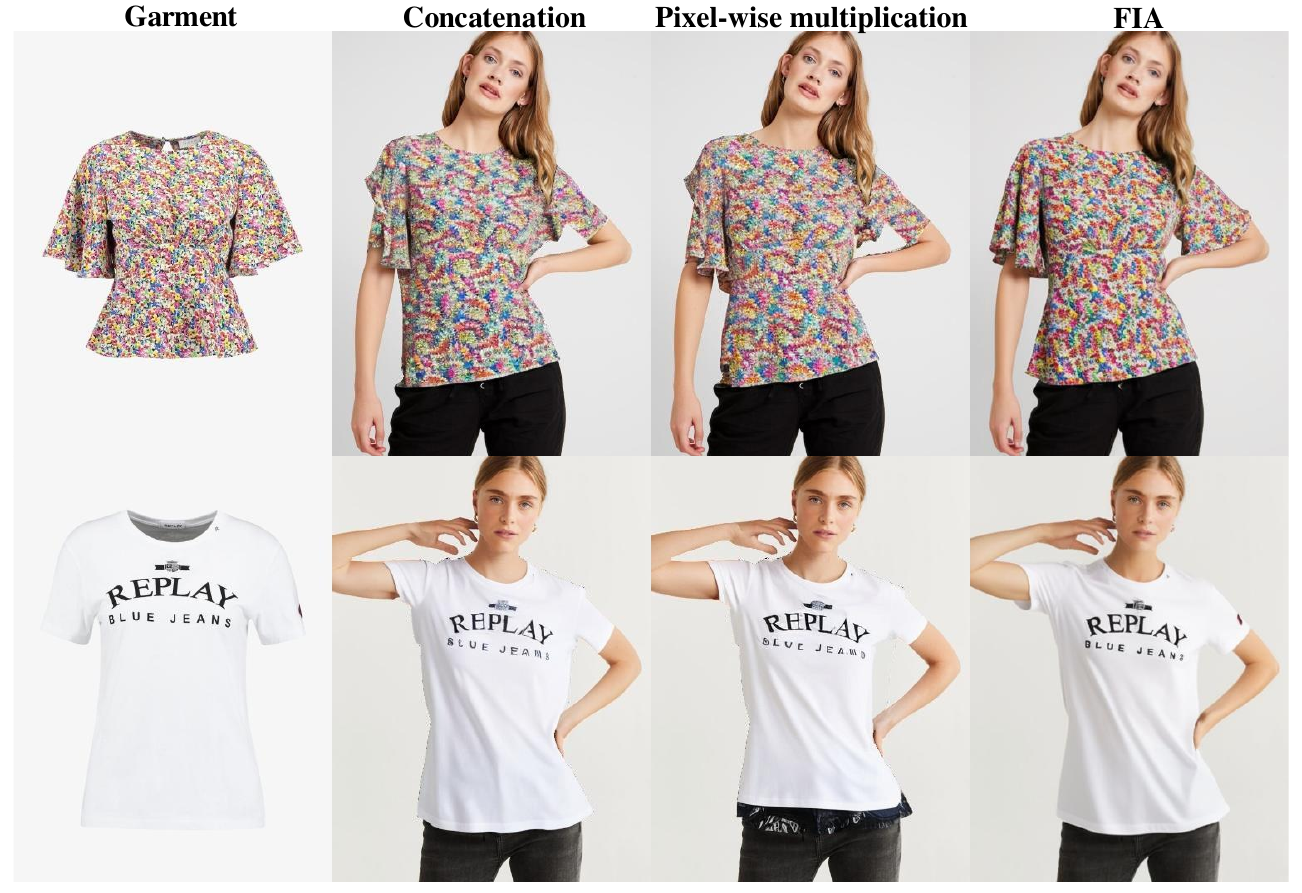}
    \vspace{-3mm}
    \caption{
    Ablation study on Flow Infused Attention (FIA).
    }
    \label{fig:warpattention_ablation}
    \vspace{-5mm}
\end{figure}

\noindent\textbf{Variants of \delong{Flow Infused Attention}.}
To demonstrate the superiority of our elaborated Flow Guider, we provide variant experiments in \Cref{fig:warpattention_ablation} and \Cref{tab:ablation_study}.
We additionally designed two modules that combine the warping module with diffusion. \delong{The "Concatenation" directly concatenates the warped garment to the input. The "Pixel-wise multiplication" warps the output features from the reference network and inputs them into the main UNet via a cross-attention mechanism. }Both methods represent explicit uses of warp flow. In \Cref{fig:warpattention_ablation}, we observed that explicitly invoking warp flow leads to two main issues:
First, generated results may lack the fine details of the original garment, especially if there is a complex pattern on the garment.
There is an obvious lack of fit between the garment and the human body, especially at the shoulders and sleeves, where the garment does not appear to be able to fully adapt to the shape of the target person's body, resulting in an incompatible fitting effect.
We speculate that such explicit guidance would rely heavily on the effectiveness of the warping module and be limited by the target mask.
In contrast, our Flow Guider module provides implicit guidance that helps mitigate these issues, resulting in more natural and visually coherent try-on results.

\subsection{In-the-wild senarios}
\begin{figure}
    \centering
    \includegraphics[width=0.95\linewidth]{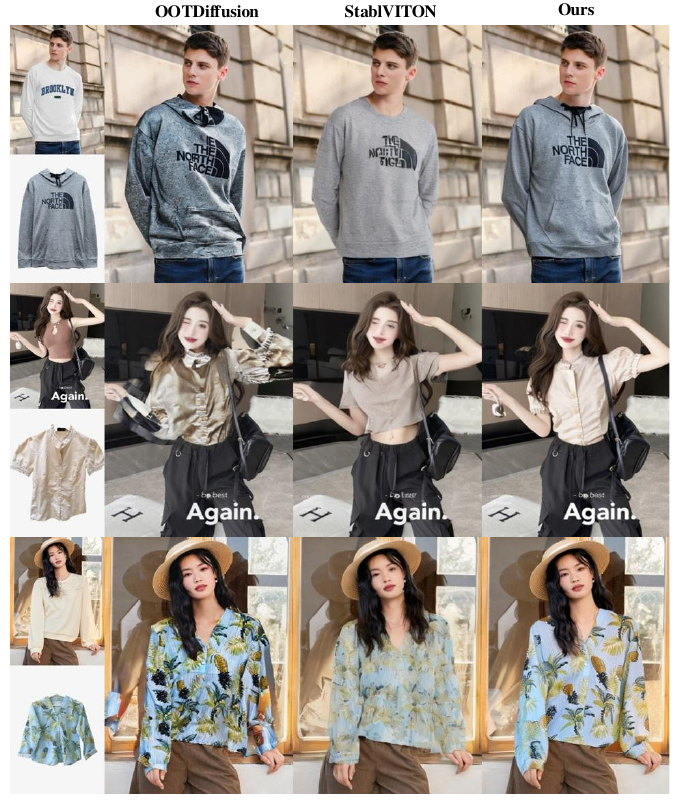}
    \vspace{-3mm}
    \caption{
    \delong{
        Qualitative comparison in the wild scenarios.
        Compared with state-of-the-art methods (OOTDiffusion, CAT-VTON).
        Our method generates more natural images that seamlessly combine background, person, and garment in complex scenarios. Zoom in for more details.
        }
    }
    \label{fig:inTheWild}
    \vspace{-3mm}
\end{figure}

Furthermore, we evaluate the wild scenarios to test the robustness and applicability of FIA-VTON in real-world conditions. As shown in \Cref{fig:inTheWild}, FIA-VTON accurately recognizes and integrates the shape of complex garments, such as off-shoulder designs, with the person. It can generate interlaced parts for complex poses, such as sitting. Additionally, it effectively completes and integrates the background with the garment in complex in-the-wild scenarios.

\section{Conclusions}
In this paper, we presented FIA-VTON, a novel approach for virtual try-on that integrates flow-guided attention and high-level semantic features to achieve realistic garment fitting and texture detail preservation. By addressing the shortcomings of previous Warp-based and Learning-based methods, FIA-VTON leverages the Flow Infused Attention module to enhance garment alignment while reducing visual artifacts. In IFA, a Flow Attention is designed to take the dense flow map itself to guide the implicit warp, and extra Spatial Attention to extract high-level semantic information, ensuring consistent garment and model integration. 
Experimental results demonstrate that our method outperforms existing approaches in both global alignment and fine-grained texture reconstruction. 
Future work may focus on extending FIA-VTON to handle more diverse clothing types and complex poses, as well as optimizing efficiency for real-time applications.

{
    \small
    \bibliographystyle{ieeenat_fullname}
    \bibliography{main}
}

\end{document}